\documentclass[runningheads]{llncs}
\usepackage{graphicx}
\usepackage{comment}
\usepackage{amsmath,amssymb} 
\usepackage{color}

\usepackage[colorlinks,linkcolor=blue,citecolor=blue]{hyperref}
\usepackage{multirow}
\usepackage{color}
\usepackage{tabularx,verbatim}
\usepackage{xspace}
\usepackage{algorithm}
\usepackage{algorithmic}
\usepackage{setspace}

\definecolor{mygray}{gray}{0.5}



\usepackage{amsmath,amsfonts,bm}

\def\eqref#1{equation~\ref{#1}}

\def\1{\bm{1}}

\newcommand{\test}{\mathcal{D_{\mathrm{test}}}}

\DeclareMathAlphabet{\mathsfit}{\encodingdefault}{\sfdefault}{m}{sl}
\SetMathAlphabet{\mathsfit}{bold}{\encodingdefault}{\sfdefault}{bx}{n}

\usepackage{cleveref}
\newcommand{\ours}{OpenMix\xspace}

\begin{document}
\pagestyle{headings}
\mainmatter
\def\ECCVSubNumber{1234}  

\title{OpenMix: Reviving Known Knowledge for \\ Discovering Novel Visual Categories \\ in An Open World}
%

\titlerunning{OpenMix}
%
\author{Zhun Zhong\inst{1}, Linchao Zhu\inst{2}, Zhiming Luo\inst{1}, Shaozi Li\inst{1}, Yi Yang\inst{2}, Nicu Sebe\inst{3}}
\authorrunning{Zhong et al.}
%
\institute{$^{1}$Xiamen University\quad $^{2}$University of Technology Sydney\quad$^{3}$ University of Trento}
\maketitle

\begin{abstract}

In this paper, we tackle the problem of discovering new classes in unlabeled visual data given labeled data from disjoint classes.
Existing methods typically first pre-train a model with labeled data, and then identify new classes in unlabeled data via unsupervised clustering. 
However, the labeled data that provide essential knowledge are often underexplored in the second step.
The challenge is that the labeled and unlabeled examples are from non-overlapping classes, which makes it difficult to build the learning relationship between them.
In this work, we introduce \ours to mix the unlabeled examples from an open set and the labeled examples from known classes, 
where their non-overlapping labels and pseudo-labels are simultaneously mixed into a joint label distribution.
\ours dynamically compounds examples in two ways.
First, we produce mixed training images by incorporating labeled examples with unlabeled examples. 
With the benefits of unique prior knowledge in novel class discovery,
the generated pseudo-labels will be more credible than the original unlabeled predictions. As a result, \ours helps to prevent the model from overfitting on unlabeled samples that may be assigned with wrong pseudo-labels. 
Second, the first way encourages the unlabeled examples with high class-probabilities to have considerable accuracy. We introduce these examples as reliable anchors and further integrate them with unlabeled samples. This enables us to generate more combinations in unlabeled examples and exploit finer object relations among the new classes.
Experiments on three classification datasets demonstrate the effectiveness of the proposed \ours, which is superior to state-of-the-art methods in novel class discovery.

\end{abstract}

\section{Introduction}

In this work, we attempt to address the new problem, called novel class discovery \cite{han2020automatically,han2019learning,hsu2017learning,hsu2019multi}, where we are given labeled data of known (old) classes and unlabeled data of novel (new) classes. It is an open set problem where classes of unlabeled data are undefined previously and annotated samples of these novel classes are not available. The goal of novel class discovery is to identify new classes in unlabeled data with the support of knowledge of old classes. To achieve this objective, existing methods \cite{han2020automatically,han2019learning,hsu2017learning,hsu2019multi} commonly follow a two-step learning strategy: 1) pre-train the model with labeled data to obtain basic discriminative ability; 2) recognize new classes in unlabeled data via unsupervised learning upon the trained model. However, the labeled data are only used to learn off-the-shell features in the first step, but are largely ignored in the second step. In this way, the model can only benefit from the off-the-shell knowledge of the labeled data, but fails to leverage the underlying relationship between the labeled and unlabeled data. In this work, we argue that the labeled data provide essential knowledge about underlying object structures and common visual patterns. However, the use of labeled data is much harder than in semi-supervised learning \cite{berthelot2019mixmatch,oliver2018realistic}, due to the labeled and unlabeled samples are from disjoint classes.

\begin{figure}[!t]
\centering
\includegraphics[width=0.96\linewidth]{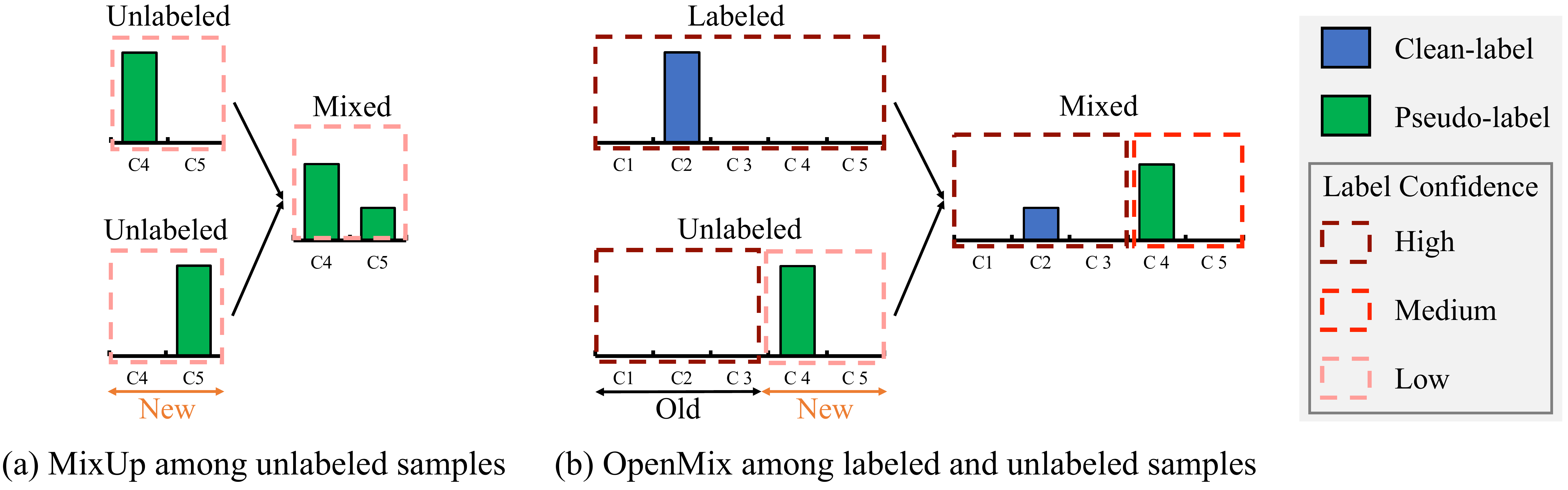}
\vspace{-.1in}
\caption{Examples of (a) directly using MixUp among unlabeled samples and (b) the proposed \ours. Due to the uncertainty of pseudo-labels of unlabeled samples, their mixed label may still have \textit{low} confidence. In \ours, the prior knowledge (area of \textit{high} confidence) leads the mixed label to have \textit{high} (exactly true) confidence in old classes and \textit{medium} (reliable) confidence in new classes.}
\label{fig:mix_comparision}
\vspace{-.12in}
\end{figure}

To this end, the question is how to effectively exploit the labeled data to promote the discovery of new classes? In this work, we try to answer this question and propose a simple but effective method, called \ours, for the open set problem considered in this paper. \ours is largely motivated by MixUp \cite{zhang2018MixUp}, which is widely used in supervised learning \cite{zhang2018MixUp,yun2019cutmix} and semi-supervised learning \cite{berthelot2019mixmatch,berthelot2020remixmatch}. However, one premise of using MixUp is that there should be labeled samples for every class of interest, which is not appropriate for our task. This is because we only have pseudo-labels for unlabeled samples of new classes, and the accuracy of these pseudo-labels can not be guaranteed. If we directly apply MixUp among unlabeled samples along with their uncertain pseudo-labels, the generated pseudo-labels will still be unreliable or even more unreliable (Fig~.\ref{fig:mix_comparision} (a)). Training with these unreliable pseudo-labels may further damage the model performance. Therefore, it is non-trivial to adopt MixUp for novel class discovery.

Instead of readily using MixUp among unlabeled samples, during the unsupervised clustering, \ours generates training samples by incorporating both labeled and unlabeled samples. \ours compounds samples in two ways. First, \ours mixes the labeled samples with unlabeled samples. Meanwhile, since the labeled and unlabeled samples belong to different label spaces, we first extend their labels/pseudo-labels to joint label distributions, and then mixed them. 
\ours leverages two prior knowledge in novel class discovery: 1) labels of labeled samples of old classes are exactly clean, and 2) labeled and unlabeled samples belong to completely different classes. These two properties encourage the pseudo-labels of mixed samples to have 1) exactly true confidence in old classes and 2) higher confidence in new classes (Fig~.\ref{fig:mix_comparision} (b)). That is, in old class set, the mixed sample only proportionally belongs to the class of its labeled counterpart. This is because the unlabeled counterpart does not belong to any old classes. On the other hand, in the new class set, the uncertainty of pseudo-label will be partially eliminated by mixing with the labeled sample. This is due to the labeled counterpart does not belong to any new classes and its label distribution in the new class set is exactly true. With the above properties, the pseudo-labels of mixed samples will be more reliable than that of their unlabeled counterparts. As a result, \ours can help preventing the model from overfitting on unlabeled samples that may be assigned wrong pseudo-labels. Second, we observe that the first way of \ours encourages the model to keep considerable classification accuracy for unlabeled samples having high class-probabilities. Therefore, we select these samples as reliable anchors of new classes and mix them with unlabeled samples for further improvement.

In summary, the contribution of this paper resides in: (1) This work proposes the \ours, which is tailor-made for effectively leveraging known knowledge in novel class discovery. \ours can prevent the model from fitting on wrong pseudo-labels, thereby consistently improving the model performance; (2) \ours enables us to explore reliable anchors from unlabeled samples, which can be used to generate diverse smooth samples of new classes towards a more discriminative model; (3) This paper presents a simple baseline for novel class discovery, which can achieve competitive results; (4) Experiments conducted on three datasets show that our approach outperforms the state-of-the-art methods by a large margin in novel class discovery.

\section{Related Work}

This work is related to novel class discovery, unsupervised clustering, transfer learning, semi-supervised learning and MixUp \cite{zhang2018MixUp}. We brieﬂy review the most representative works and discuss the relationship with them.

\textbf{Novel Class Discovery} is a recent task aiming at recognizing novel classes in unlabeled data. Different from the traditional unsupervised learning, this task also provides labeled data of other classes. Existing methods usually use the labeled data for model initialization and perform unsupervised clustering on unlabeled data. In \cite{hsu2017learning} and \cite{hsu2019multi}, a Constrained Clustering Network (CCN) is proposed. CCN first trains a binary-classification model on labeled data to estimate pair-wise similarity of images. Then, a clustering model is trained on unlabeled data by using the prediction of the binary-classification model as supervision. The difference between \cite{hsu2017learning} and \cite{hsu2019multi} is the loss function used in CCN. 
Han \textit{et al.} \cite{han2019learning} first pre-train the model on labeled data by cross-entropy loss and then implement clustering on unlabeled data by DEC \cite{xie2016DEC}. Latter, Han \textit{et al.} \cite{han2020automatically} propose employing rank statistics to estimate the pairwise similarity of images. The pairwise pseudo-labels are used to achieve unsupervised clustering on the unlabelled data. Except \cite{han2020automatically}, none of the above methods use the labeled data during the stage of unsupervised clustering. In \cite{han2020automatically}, the labeled data are mainly used to keep the model accuracy on old classes. By contrast, our goal is improving the accuracy on new classes with the labeled data.

\textbf{Unsupervised Clustering} focuses on automatically dividing unlabeled data. Many classic methods \cite{macqueen1967_kmeans,zelnik2005self} and deep learning methods \cite{xie2016DEC,ghasedi2017deep,yang2017towards} have been proposed. Unlike novel class discovery, there is no prior knowledge provided (\textit{e.g.}, labeled data) for unsupervised clustering. In such a context, there may be multiple criteria for most datasets, such as color, shape, and other attributes, so that the clustering results may not fit the expectation. In contrast, the labeled data in novel class discovery provide useful knowledge and can guide us to learn clustering models that match the clustering criteria of labeled data.

\textbf{Transfer Learning} \cite{pan2009survey,weiss2016survey,tan2018survey} aims to transfer the knowledge of a labeled dataset to another dataset. Generally, the classes of the new (target) dataset are different from the previous (source) one. In transfer learning, both the source and target data are labeled. Instead, the target data are unlabeled in novel class discovery, leading this task to be more difficult.

\textbf{Semi-Supervised Learning} \cite{tarvainen2017mean,oliver2018realistic,berthelot2019mixmatch,berthelot2020remixmatch} is designed to training a model on a partially labeled dataset. Novel class discovery is similar to this task in that both tasks are provided with labeled and unlabeled samples. The difference is that the labeled and unlabeled samples share the same class set in semi-supervised learning. However, the classes of labeled and unlabeled samples are completely different in novel class discovery.

\textbf{MixUp} \cite{zhang2018MixUp} has been utilized successfully in supervised learning \cite{yun2019cutmix,zhang2018MixUp} and semi-supervised learning \cite{berthelot2020remixmatch,berthelot2019mixmatch}. Unlike existing MixUp-based methods, we apply MixUp to effectively leverage labeled data of known classes for novel class discovery. In addition, existing MixUp-based methods assume that there are some clean labels for every class of interest, which is an important precondition. However, in novel class discovery, the labels of new classes are not available, causing MixUp not to be directly applicable without careful design.

\section{Our Method}

\begin{figure}[!t]
\centering
\includegraphics[width=0.999\linewidth]{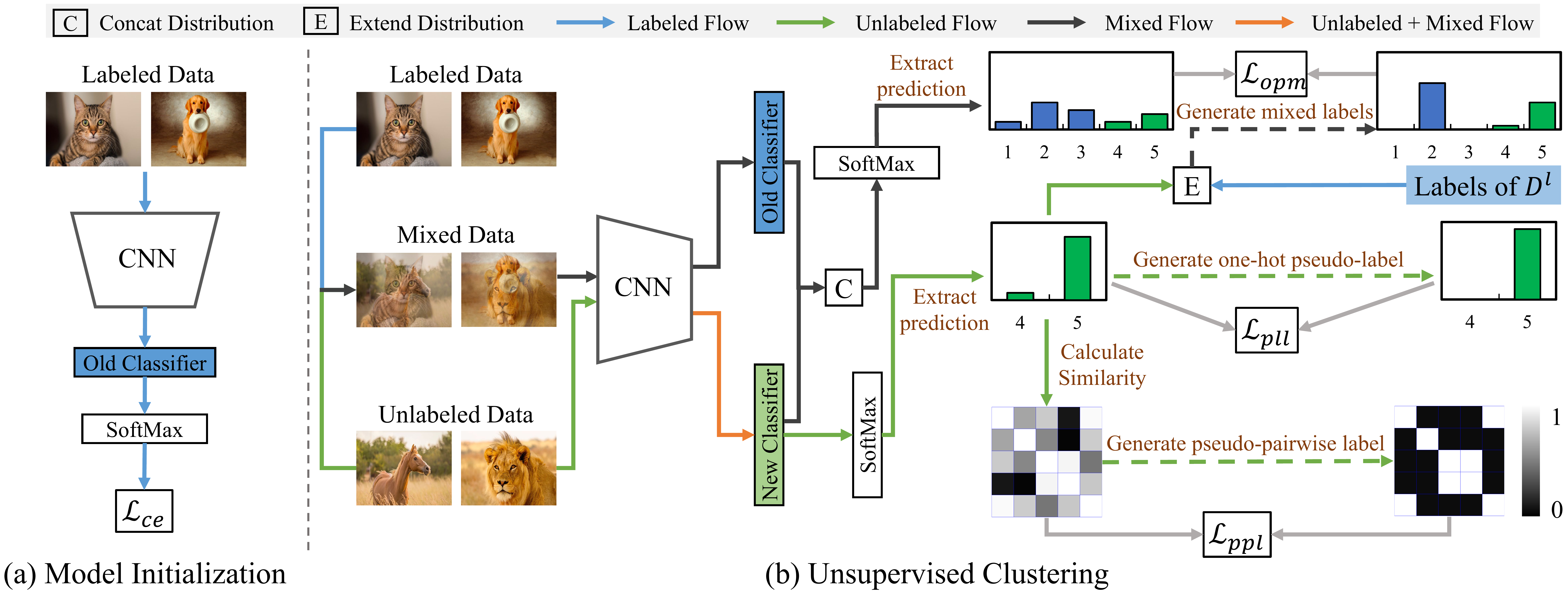}
\vspace{-.2in}
\caption{The pipeline of the proposed method. (a) We first initialize the model on the labeled data ($\mathcal{L}_{ce}$). (b) Then, we learn the unsupervised clustering model for discovering new classes in unlabeled data, by pseudo-pair learning ($\mathcal{L}_{ppl}$), pseudo-label learning ($\mathcal{L}_{pll}$) and learning with the proposed \ours ($\mathcal{L}_{opm}$).}
\label{fig:framework}
\vspace{-.1in}
\end{figure}

\textbf{Preliminary} In novel class discovery, we are provided with labeled data $D^l$ = \{$X^l, Y^l$\} and unlabeled data $D^u$ = \{$X^u$\}. The numbers of samples are $N^l$ in $D^l$ and $N^u$ in $D^u$, respectively. Each labeled image $x_i^l$ has a label $y_i^l$, where $y_i^l \in \{1, 2, ..., C^l\}$ and $C^l$ is the number of classes of $D^l$. Following \cite{han2020automatically}, we suppose the number of classes of $D^u$ is prior knowledge, which is defined as $C^u$. The classes of $D^l$ and $D^u$ are disjoint. We define the classes as old classes and new classes for $D^l$ and $D^u$, respectively. The goal of novel class discovery is to leverage the knowledge of $D^l$ to identify the classes in $D^u$.

In this paper, we try to achieve this goal by learning a model constructed by a convolutional neural network (CNN) and two classifiers. These two classifiers, called old classifier and new classifier, are used to recognize samples from old classes and new classes, respectively. The framework of our method is illustrated in Fig.~\ref{fig:framework}. Next, we will present the baseline model for novel class discovery.

\subsection{Baseline}
\label{section:3:Baseline}

In this work, we follow the two-stage learning strategy to design the baseline. In the first stage, we utilize the labeled data to train the CNN and the old classifier, which can provide basic discriminative representations for images and accurately classify samples of old classes. In the second stage, we learn an unsupervised clustering model on the unlabeled data by pseudo-pair learning and pseudo-label learning, enabling us to identify samples of new classes.

\noindent\textbf{Stage 1: Model Initialization}
Given the labeled data $D^l$ = \{$X^l, Y^l$\}, we are able to train the model in a supervised way. Specifically, the model is trained with the cross-entropy loss, as done in the traditional supervised classification \cite{krizhevsky2012alexnet,resnet}. The loss function is formulated as,
\begin{eqnarray}
    \begin{array}{l}
   \mathcal{L}_{ce} = - \frac{1}{n^l} \sum\limits_{i=1}^{n^l}  \log[{\rm SoftMax}(z_i^l)]^\top \cdot \hat{y}^l_i,
   \label{cross-entropy-loss}
   \end{array}
\end{eqnarray}
where $n^l$ is the number of labeled training samples in a mini-batch, $z_i^l \in \mathbb{R}^{C^l}$ is the output of old classifier, and $\hat{y}^l_i \in \mathbb{R}^{C^l}$ is the one-hot label converted by $y^l_i$.

\noindent \textbf{Stage 2: Unsupervised Clustering}

\noindent\textbf{Pseudo-Pair Learning}.
Given the model pre-trained on the labeled data, we additionally add a classifier layer of $C^u$ new classes on the head of CNN. We then focus on the second stage, \textit{i.e.}, unsupervised clustering in unlabeled data. To achieve this goal, we first explore the relationship between two images for model training. Inspired by DAC \cite{Chang_2017_ICCV} and DCCM \cite{wu2019deep}, we argue that the relation of pairwise images is binary. In other words, each pair of images should be either of the same class or different classes. In light of this, we convert the unsupervised clustering problem to a binary classification one, aiming to distinguish whether a pair of images belong to the same class.

Similar to DAC \cite{Chang_2017_ICCV} and DCCM \cite{wu2019deep}, we first obtain the outputs of the new classifier for input unlabeled samples and compute their cosine similarity matrix $\mathcal{S} \in \mathbb{R}^{n^u \times n^u}$, where
\begin{eqnarray}
    \begin{array}{l}
   \mathcal{S}_{i, j} = \frac{(\hat{z}_i^{u})^\top . \hat{z}_j^u}{\|\hat{z}_i^{u}\|_2  \|\hat{z}_j^u\|_2},~ \hat{z}_i^{u} = {\rm SoftMax}(z_i^u).
   \label{matrix}
   \end{array}
\end{eqnarray}
$z_i^u \in \mathbb{R}^{C^u}$ is the output of the new classifier. $n^u$ denotes the number of unlabeled training images in a mini-batch. We then estimate the pseudo-pairwise labels $\mathcal{W}$ by setting a threshold $\theta_1$ on $\mathcal{S}$, where
\begin{eqnarray}
  \mathcal{W}_{i, j} =
  \begin{cases}
   0, & \mathcal{S}_{i, j} < \theta_1 \\ 
    1, & \mathcal{S}_{i, j} \geq \theta_1 
  \end{cases}.
\label{binary-label}
\end{eqnarray}
By doing so, two images are defined as a positive pair if the cosine similarity of them is larger than $\theta_1$, otherwise they are a negative pair. Given this pairwise supervision, we train the model with a binary cross-entropy loss, formulated as,
\begin{eqnarray}
   \begin{array}{r}
   \mathcal{L}_{ppl} = - \frac{1}{(n^u)^2} \sum\limits_{i, j} \big (\mathcal{W}_{i, j} \log \mathcal{S}_{i, j} + (1 - \mathcal{W}_{i, j}) \log (1 - \mathcal{S}_{i, j})\big ), \\ \forall i, j \in \{1, 2, ..., n^u\}.
   \label{ppl-loss}
   \end{array}
\end{eqnarray}

\noindent\textbf{Pseudo-Label Learning}.
According to the proof given by DAC \cite{Chang_2017_ICCV} and DCCM \cite{wu2019deep}, the constraint $\mathcal{W}_{i, j}$ between images $x_i^u$ and $x_j^u$ defined in Eq.~\ref{ppl-loss} can bring the following clustering property:
\noindent \textit{If the optimal solution of Eq.~\ref{ppl-loss} is achieved, $\forall i, j$, $\hat{z}^u \in \mathbb{R}^{C^u}$, $\hat{z}_i^u = \hat{z}_j^u \Leftrightarrow$ $\mathcal{W}_{i, j}$ = 1, and, $\hat{z}_i^u \ne \hat{z}_j^u \Leftrightarrow$ $\mathcal{W}_{i, j}$ = 0}.

\noindent This property denotes that the predictions of the optimal new classifier, $\hat{z}^u$, are exactly $C^u$-diverse one-hot vectors. In other words, the unlabeled data $D^u$ can be automatically divided into $C^u$ partitions.

Based on this property, for unlabeled samples, we reformulate their predictions output by the new classifier to one-hot pseudo-labels, which can be used to further improve the model performance. The one-hot pseudo-label $\hat{y}_i^u$ of an unlabeled image $x_i^u$ is generated by setting a threshold $\theta_2$ on $\hat{z}_i^u$, where
\begin{eqnarray}
  \hat{y}_i^u[j] =
  \begin{cases}
   0, & \hat{z}_i^u[j] < \theta_2 \\ 
    1, & \hat{z}_i^u[j] \geq \theta_2 
  \end{cases}.
\label{pseduo-label}
\end{eqnarray}

In the pseudo-label learning, we only train the model with the unlabeled samples that are assigned with one-hot pseudo-labels, \textit{i.e.}, ${\rm Max} (\hat{y}^u) = 1$. Given the one-hot pseudo-labels for unlabeled samples, we are able to train the model with cross-entropy loss, formulated as,
\begin{eqnarray}
   \begin{array}{l}
   \mathcal{L}_{pll} = - \frac{1}{\hat{n}^u} \sum\limits_{i}^{} \log ({\hat{z}_i^u}){^{\top}} \cdot \hat{y}_i^u,~~\forall i \in \{{\rm Max} (\hat{y}_i^u) = 1\},
   \label{pseudo-entropy-loss}
   \end{array}
\end{eqnarray}
where $\hat{n}^u$ is the number of unlabeled samples that are assigned with one-hot pseudo-labels in a mini-batch.

\noindent \textbf{Combination of Two Losses}.
By jointly considering the pseudo-pair learning and pseudo-label learning, the unsupervised clustering loss is expressed as,
\begin{eqnarray}
    \begin{array}{l}
   \mathcal{L}_{uc} = \mathcal{L}_{ppl} + \lambda_1 \mathcal{L}_{pll},
   \label{uc-loss}
   \end{array}
\end{eqnarray}
where $\lambda_1$ is the hyper-parameter that controls the importance of pseudo-label learning. To this end, we have presented our baseline for novel class discovery.

\subsection{\ours}
\label{section:3:OpenMix}

In the baseline presented in Section~\ref{section:3:Baseline}, the labeled data only play the role of model initialization. However, there is no utilization of labeled data in the second unsupervised clustering stage. In this paper, we argue that the labeled data can provide important knowledge for improving the unsupervised clustering. In this section, we propose the \ours for effectively leveraging the labeled data $D^l$ during the unsupervised clustering in unlabeled data $D^u$. \ours is easy to implement. In a nutshell, during unsupervised clustering, \ours additionally compounds examples in two ways: 1) mix unlabeled examples with labeled samples; and 2) mix unlabeled examples with reliable anchors.

\noindent\textbf{Mix with Labeled Examples}. In the first way, \ours mixes the labeled samples with unlabeled samples, as well as their labels with pseudo-labels. Taking the prior knowledge that labeled samples and unlabeled samples belong to completely different classes, we first extend the label distributions of the labeled samples and unlabeled samples to the same size. Specifically, we concatenate $\hat{y}^l$ with a $C^u$-dim zeros-vector while $\hat{z}^u$ with a $C^l$-dim zeros-vector. The extended labels/pseudo-labels are represented by $\bar{y}^l$ for labeled samples and $\bar{y}^u$ for unlabeled samples, respectively. We then generate virtual sample with MixUp~\cite{zhang2018MixUp},
\begin{eqnarray}
   \begin{array}{l}
    \eta \thicksim {\rm Beta}(\epsilon, \epsilon), ~~~
    {\eta}^* = {\rm Max}(\eta, 1-\eta), \\
     m = {\eta}^* x^l + (1- {\eta}^*) x^u,  ~~~
    v = {\eta}^* \bar{y}^l + (1- {\eta}^*) \bar{y}^u,
   \label{eq:MixUp}
   \end{array}
\end{eqnarray}
where $\epsilon$ is a hyper-parameter and $\eta \in [0, 1]$. $m$ is the generated sample and $v$ is the pseudo-label of $m$. The second constraint in Eq.~\ref{eq:MixUp} ensures that the generated sample $m$ is closer to $x^l$ than $x^u$. This can alleviate the negative impact caused by unreliable pseudo-labels of unlabeled samples.

As shown in Fig.~\ref{fig:mix_comparision} (b), the mixed sample has exactly true confidence in the old classes and medium confidence in the new classes. This is benefited from the prior knowledge, \textit{i.e.}, the label of labeled sample is exactly true, and, the classes of labeled and unlabeled samples are completely different. Therefore, by mixing labeled samples with unlabeled samples through \ours, the pseudo-labels of mixed samples will be more reliable than that of their unlabeled counterparts. Learning with the mixed samples can help prevent the model from overfitting on unlabeled samples that are assigned with wrong pseudo-labels.

\noindent\textbf{Mix with Reliable Anchors}. By training with samples generated by the first way, we observe that the model keeps considerable accuracy for unlabeled samples that are predicted with high class-probabilities (${\rm Max} (\hat{z}^u) \ge \theta_2$). Based on this observation, in second way, we further select the unlabeled samples that have high class-probabilities as reliable anchors. Then, we mix the anchors with unlabeled samples through \ours. Specifically, we perform this operation by replacing the labeled sample $x^l$ with a reliable anchor in Eq.~\ref{eq:MixUp}.

\noindent \textbf{Loss of \ours}. Given mixed samples $\mathcal{M}$ and their pseudo-labels $\mathcal{V}$, we apply L2-norm loss to train the model, defined as,
\begin{eqnarray}
 \begin{array}{l}
  \mathcal{L}_{opm} = \frac{1}{|\mathcal{M}|} \sum\limits_{i \in \mathcal{M}} \frac{1}{C^l+C^u} \| v_i - {\rm SoftMax}(z_i^m) \|_2,
  \label{eq:loss-opm}
 \end{array}
\end{eqnarray}
where $|\mathcal{M}|$ denotes the number of samples in $\mathcal{M}$. $z_i^m$ indicates the concentrated outputs of the old and new classifiers. Specifically, we forward $m_i$ to the model and extract the outputs of the old and new classifiers, which are represented as $z^l_i$ and $z^u_i$, respectively. $z_i^m$ is then obtained by concentrating $z^l_i$ and $z^u_i$. 

\subsection{Overall Loss}
\label{section:3:overall}

By combining the baseline and the proposed \ours, the overall loss of our method is expressed as,
\begin{eqnarray}
    \begin{array}{l}
   \mathcal{L}_{all} = \mathcal{L}_{uc} + \lambda_2 \mathcal{L}_{opm},
   \label{overall-loss}
   \end{array}
\end{eqnarray}
where $\lambda_2$ is the hyper-parameter that balances the weight of \ours.

\subsection{Discussion}
In this section, we analyze the label reliability of mixed samples generated by MixUp and \ours over their unlabeled counterparts. Given two unlabeled samples $x_a^u$ and $x_b^u$, we denote their ground-truth labels and pseudo-labels as \{$Y_a$, $Y_b$\} and \{$\hat{Y}_a$, $\hat{Y}_b$\}, respectively. The label error of a sample is calculated by the L1-norm loss,
\begin{eqnarray}
 \begin{array}{l}
   E(Y, \hat{Y}) = \sum\limits_{i}^{N^u} \big|y[i] - \hat{y}[i] \big|,
   \label{eq:label-error}
  \end{array}
\end{eqnarray}
where $y[i]$ and $\hat{y}[i]$ indicate the probability of the $i$-th class.

We first directly apply MixUp among $x_a^u$ and $x_b^u$. The ground-truth label and pseudo-label of the mixed sample are represented as $M_{ab}$ and $\hat{M}_{ab}$, respectively. According to Eq.~\ref{eq:label-error}, the label error of the mixed sample is written as,
\begin{equation}
  \begin{array}{l}
   E(M_{ab}, \hat{M}_{ab}) = \sum\limits_{i}^{N^u} \big |\eta(y_a[i] - \hat{y}_a[i]) + (1- \eta)(y_b[i] - \hat{y}_b[i]) \big |,
   \label{eq:difference:mixup}
  \end{array}
\end{equation}
where $\eta$ is the mixing weight. The difference of label error between $x_b^u$ and the mixed sample of MixUp is calculated as,
\begin{equation}
\begin{array}{r}
\noindent E(Y_b, \hat{Y}_b)\!-\!E(M_{ab}, \hat{M}_{ab}) \!=\! \sum\limits_{i}^{N^u} \big|y_b[i] \!-\! \hat{y}_b[i]\big| \!-\! \big|\eta(y_a[i] \!-\! \hat{y}_a[i]) \!+\! (1 \!-\! \eta)(y_b[i] \!-\! \hat{y}_b[i])\big|.
   \label{eq:difference:mixup2original}
 \end{array}
\end{equation}
We can use proof by contradiction to verify that Eq.~\ref{eq:difference:mixup2original} is not always larger than 0. For example, when $N^u = 1$, $\eta = 0.6$, $y_a[1] - \hat{y}_a[1] = - 0.8$, and $y_b[1] - \hat{y}_b[1] = 0.2$, we can obtain that Eq.~\ref{eq:difference:mixup2original} = $-0.2$. Thus, there is no guarantee that the label error of sample generated by MixUp can be lower than that of $x_b^u$. Moreover, the label errors of $x_a^u$ and $x_b^u$ may be accumulated during mixing, leading the label error of the mixed sample to be larger.

Next, we implement \ours among a labeled sample $x_c^l$ and the unlabeled sample $x_b^u$. We first extend their label distributions to the size of $C^l+C^u$ as described in Section~\ref{section:3:OpenMix}. Then, we mix $x_c^l$ with $x_b^u$. The ground-truth label and pseudo-label of the mixed sample of \ours are represented as $P_{cb}$ and $\hat{P}_{cb}$, respectively. The label error of the mixed sample of \ours is computed as,
\begin{eqnarray}
    \begin{array}{l}
   E(P_{cb} - \hat{P}_{cb}) = \sum\limits_{i}^{N^l+N^u}   \big|\eta(y_c[i] - \hat{y}_c[i]) + (1- \eta)(y_b[i] - \hat{y}_b[i])\big|,
   \label{eq:difference:opm}
   \end{array}
\end{eqnarray}
where $i \in [1, N^u]$ represents the new class and $i \in [N^u+1, N^l+N^u]$ indicates old class. Since the label of $x_c^l$ is available during training, the pseudo-label for $x_c^l$ is the same as the ground-truth label. Therefore, the first term in Eq.~\ref{eq:difference:opm} is 0. In addition, we added zeros-vector to old classes for the ground-truth label and pseudo-label of $x_b^u$, thus $y_b[i] = \hat{y}_b[i] = 0, \forall i \in [N^u+1, N^u+N^l]$. Taking these two conditions, Eq.~\ref{eq:difference:opm} can be reformulated as,
\begin{eqnarray}
    \begin{array}{l}
   E(P_{cb}, \hat{P}_{cb}) = \sum\limits_{i}^{N^u}  \big|(1- \eta)(y_b[i] - \hat{y}_b[i]) \big|.
   \label{eq:difference:opm-re}
   \end{array}
\end{eqnarray}
The difference of label error between $x_b^u$ and the mixed sample of \ours is calculated as,
\begin{eqnarray}
    \begin{array}{l}
   E(Y_b, \hat{Y}_b) - E(P_{cb}, \hat{P}_{cb}) = \sum\limits_{i}^{N^u} \eta \big|y_b[i] - \hat{y}_b[i]\big| \ge 0.
   \label{eq:difference:opm2original}
   \end{array}
\end{eqnarray}
Due to $\eta \in [0, 1]$, E($P_{cb}, \hat{P}_{cb}$) is always lower than E($Y_b, \hat{Y}_b$), except when the pseudo-label of $x_b^u$ is exactly accurate. In other words, the sample generated by \ours always has higher label reliability of pseudo-label than its unlabeled counterpart.

\section{Experiments}

\subsection{Datasets and Settings}

\noindent \textbf{Datasets.} In this paper, we evaluate our method on three image classification benchmarks, including CIFAR-10 \cite{krizhevsky2009learning}, CIFAR-100 \cite{krizhevsky2009learning} and ImageNet \cite{deng2009imagenet}. Following \cite{han2020automatically}, we conduct the experiment on the setting where the number of classes in unlabeled data is known.
\textbf{CIFAR-10} \cite{krizhevsky2009learning} includes 50,000 training images and 10,000 test images from 10 classes. Each image has a size of $32 \times 32$. For novel class discovery, we regard the samples of the first five classes (\textit{i.e.,} airplane, automobile, bird, cat and deer) as labeled data while the remaining samples as unlabeled data.
\textbf{CIFAR-100} \cite{krizhevsky2009learning} is similar to CIFAR-10, except that samples of CIFAR-100 are drawn from 100 classes. We regard the samples of first 80 classes as labeled data, the samples of last 10 classes as unlabeled data, and the remaining samples as validation data.
\textbf{ImageNet} \cite{deng2009imagenet} contains 1.28 million training images from 1,000 classes. Following \cite{vinyals2016matching,hsu2017learning,hsu2019multi}, we divide the ImageNet into two splits, which contain 822 and 118 classes, respectively. We use the 822-class split as the labeled set. Three 30-class subsets randomly sampled from the 118-class split are used as unlabeled sets. 

\noindent \textbf{Evaluation.} We employ the clustering accuracy (ACC) and normalized mutual information (NMI) \cite{strehl2002cluster} as the metrics to evaluate the clustering performance of new classes. Both metrics are ranged from 0 to 1. Higher scores mean better performance. For CIFAR-10 and CIFAR-100, we show the average results of 10 runs. For ImageNet, results averaged in three different subsets are reported.

\subsection{Implementation Details}

For a fair comparison, we follow \cite{hsu2017learning,hsu2019multi,han2019learning} and use the 6-layer VGG \cite{vgg} like architecture / ResNet-18 \cite{resnet} network for CIFAR-100 / \{CIFAR-10, ImageNet\}.
For all three datasets, we pre-train the CNN and old classifier on the labeled data with common practice of supervised image classification \cite{resnet}.
Given the pre-trained model, we add a new classifier on the head of the CNN and train the clustering model. Specifically, we use RMSprop as the optimizer to train the model. The learning rate is kept to 0.0001 throughout the training process. We train the model for a total of 200/400/100 epochs for CIFAR-10/CIFAR-100/ImageNet. The batch sizes of unlabeled data and mixed data are both set to 64. During training, we fix the CNN and only train the new classifier at the first 60/50/100 epochs for CIFAR-10/CIFAR-100/ImageNet. Then, we train the whole model in the remaining epochs. For \ours, we inject the two ways of \ours at the 2-th epoch and 5-th epoch, respectively.
For all experiments, we set $\theta_1 = 0.95$, $\theta_2 = 0.9$, $\lambda_1 = 5$, $\lambda_2 = 1000$, and $\epsilon = 1$, which can consistently achieve well performance across datasets.

\begin{table}[!t]
\caption{Ablation study on CIFAR-10 and CIFAR-100. \textbf{PPL}: pseudo-pair learning, \textbf{PLL}: pseudo-label learning, \textbf{MixUp}: original MixUp \cite{zhang2018MixUp}, \textbf{Extend}: extend the label distribution, \textbf{\ours}: the proposed \ours; \textbf{L}: labeled samples, \textbf{U}: unlabeled samples, \textbf{A}: anchors selected from unlabeled samples.}
\vspace{-.2in}
\footnotesize
\begin{center}
\newcolumntype{C}{>{\centering\arraybackslash}X}%
\newcolumntype{R}{>{\raggedleft\arraybackslash}X}%
\begin{tabularx}{0.75\linewidth}{l|CC|cC}
\hline
\multirow{2}{*}{Method} & \multicolumn{2}{c|}{CIFAR-10} & \multicolumn{2}{c}{CIFAR-100}\\
 &  ACC & NMI & ACC & NMI\\
\hline
Baseline & 90.9\% & 0.787& 81.2\% &0.689\\ 
Baseline w/o PPL & 70.8\% & 0.691 & 23.9\% & 0.094\\
Baseline w/o PLL & 90.0\% & 0.767 & 77.2\% & 0.638\\
\hline
Basel. + MixUp (U) & 80.2\% & 0.575& 78.2\% & 0.683 \\
Basel. + MixUp (A) & 79.4\% & 0.553 & 77.6\% & 0.649 \\
Basel. + Extend (L) & 90.7\% & 0.785 & 81.8\% & 0.709 \\
Basel. + Extend (U) & 90.8\% & 0.789& 81.5\% & 0.702 \\
Basel. + Extend (L+U) & 91.4\% & 0.781 & 81.9\% & 0.708\\
\hline
Basel. + \ours w/o A & 93.3\% & 0.828 & 84.5\% & 0.733\\
Basel. + \ours w/o L & 75.2\% & 0.486& 81.6\% & 0.690 \\
Basel. + \ours & \bf 95.3\% & \bf 0.879 & \bf 87.2\%& \bf 0.754\\
\hline
\end{tabularx}
\end{center}
\vspace{-.1in}
\label{tabel:ablation}
\end{table}

\subsection{Evaluation}
\textbf{Ablation study on baseline}. In Table~\ref{tabel:ablation}, we first investigate the two components in the baseline model, \textit{i.e.}, pseudo-pair learning (PPL) and  pseudo-label learning (PLL). For comparison, we remove one of them from the baseline model and train the model. As shown in Table~\ref{tabel:ablation}, each component is contributed to improve the performance. Among them, pseudo-pair learning is most important to novel class discovery. Without pseudo-pair learning, the clustering accuracy is significantly reduced from 90.9\% to 70.8\% on CIFAR-10, and, the model fails to converge on CIFAR-100 (ACC=23.9\%). 
When removing pseudo-label learning from the baseline model, the performance will not be greatly affected in CIFAR-10 but will be largely reduced in CIFAR-100. For example, the results of the model trained without pseudo-label learning (``baseline w/o PLL'') are lower than the baseline by 4\% in ACC and by 0.051 in NMI. Considering these two components together achieves the best results, which demonstrates the mutual benefit of them.

\textbf{Ablation study on \ours}. To validate the effectiveness of the proposed \ours, we compare \ours with the following variants:
\begin{itemize}
    \item MixUp (U) / MixUp (A): directly apply MixUp \cite{zhang2018MixUp} among unlabeled samples / anchors selected from unlabeled samples.
    \item  Extend (L) /  Extend (U) /  Extend (L+U): extend the label distributions of labeled samples / unlabeled samples / labeled + unlabeled samples to the size of $C^l + C^u$, and use them to train the model with cross-entropy loss.
    \item \ours w/o L: apply \ours without mixing labeled samples with unlabeled samples; \ours w/o A: apply \ours without mixing anchors with unlabeled samples.
\end{itemize}
From the comparisons in Table~\ref{tabel:ablation}, we obtain the following observations:

\begin{enumerate}
    \item Without any modification, directly applying MixUp \cite{zhang2018MixUp} among unlabeled samples fails to achieve improvement. Both ``Basel. + MixUp (U)'' and ``Basel. + MixUp (A)'' reduce the results of the baseline model. The main reason is that labeled samples of new classes are not available in novel classes discovery and pseudo-labels of unlabeled samples are unreliable. Mixing samples that are assigned with unreliable pseudo-labels may still generate unreliable ones. This may lead the model overfitting on samples that are assigned with wrong pseudo-labels, thus harming the model performance.
    
    \item Extending the label distributions to the distributions that include both old and new classes may slightly improve the performance. The improvement is benefited from the weak supervision that labeled data and unlabeled data belong to disjoint classes. For example, ``Basel. + Extend (L+U)'' improves the clustering accuracy of the baseline from 81.2\% to 81.9\% for CIFAR-100. 
    
    \item Applying \ours between labeled samples and unlabeled samples (``Basel. + \ours w/o A'') can consistently improve the performance of the baseline. For instance, ``Basel. + \ours w/o A'' achieves 2.4\% and 3.3\% improvement in clustering accuracy compared to the baseline for CIFAR-10 and CIFAR-100, respectively. In addition, applying \ours between labeled samples and unlabeled samples is an essential step in our method. When applying \ours only between anchors and unlabeled samples (``Basel. + \ours w/o L''), the improvement is very limited on CIFAR-100, or even negative on CIFAR-10.
    
    \item Additionally performing \ours between anchors and unlabeled samples based on ``Basel. + \ours w/o A'' can further increase the performance. For example,  ``Basel. + \ours'' surpasses ``\ours w/o A'' by 2\% and 2.7\% in clustering accuracy for CIFAR-10 and CIFAR-100, respectively. With the benefit of ``Basel. + \ours w/o A'', we are able to select reliable anchors from unlabeled samples and utilize them to dominate the mixing process for further improvement.
\end{enumerate}

In conclusion, the first two observations indicate that directly using MixUp \cite{zhang2018MixUp} or extending label distribution fails to effectively improve the performance for novel class discovery. On the other hand, the latter two observations demonstrate the effectiveness of the proposed \ours for novel class discovery. In addition, it should be emphasized that the proposed \ours is not simply implementing MixUp in novel class discovery. Instead, we explicitly consider the prior knowledge and carefully design a method for mixing examples from disjoint classes.

\begin{figure}[!t]
\centering
\includegraphics[width=0.9\linewidth]{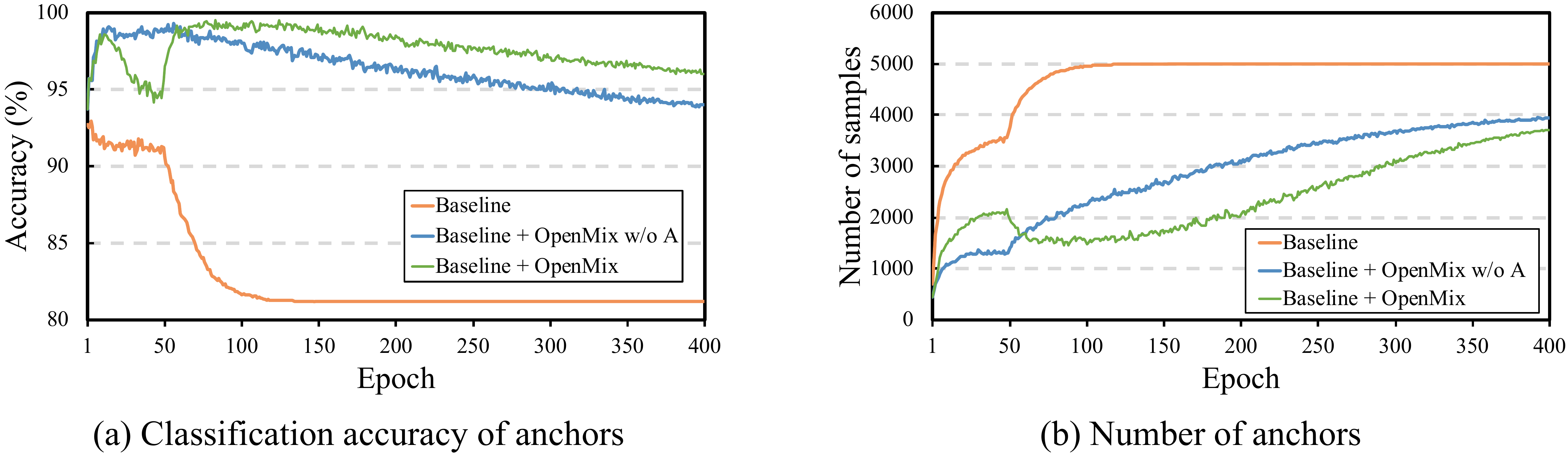}
\vspace{-.1in}
\caption{Evolution of (a) classification accuracy of selected anchors and (b) the number of selected anchors throughout training. Experiment is conducted on CIFAR-100.}
\label{fig:acc4labels}
\vspace{-.1in}
\end{figure}

\textbf{Investigation of selected anchors.} In Fig.~\ref{fig:acc4labels}, we evaluate the classification accuracy of selected anchors and the number of selected anchors throughout the training process. It is clear that the model trained with using \ours among labeled data and unlabeled data (``Baseline + \ours w/o A'') achieves much higher accuracy than the baseline model. In addition, the accuracy of ``Baseline + \ours w/o A'' can always be maintained above 95\%. The full version of \ours, which additionally mixes selected anchors with unlabeled samples, will slightly reduce the accuracy after 20 epochs. 
The reduction in accuracy is mainly caused by introducing more combinations of unlabeled samples. Due to the CNN is fixed at the first 50 epochs, learning with more combinations may lead the new classifier to overfit on unreliable samples.
However, the accuracy of \ours will quickly increase after 50 epochs and will be higher than that of ``Baseline + \ours w/o A'' after 80 epochs. From Fig.~\ref{fig:acc4labels} (b), we can observe that the numbers of selected anchors of two \ours-based methods are consistently less than that of the baseline model. The above observations indicate that the proposed \ours can prevent the model from overfitting on wrong pseudo-labels and ensures the model to train with cleaner samples.

\begin{figure}[!t]
\centering
\includegraphics[width=0.9\linewidth]{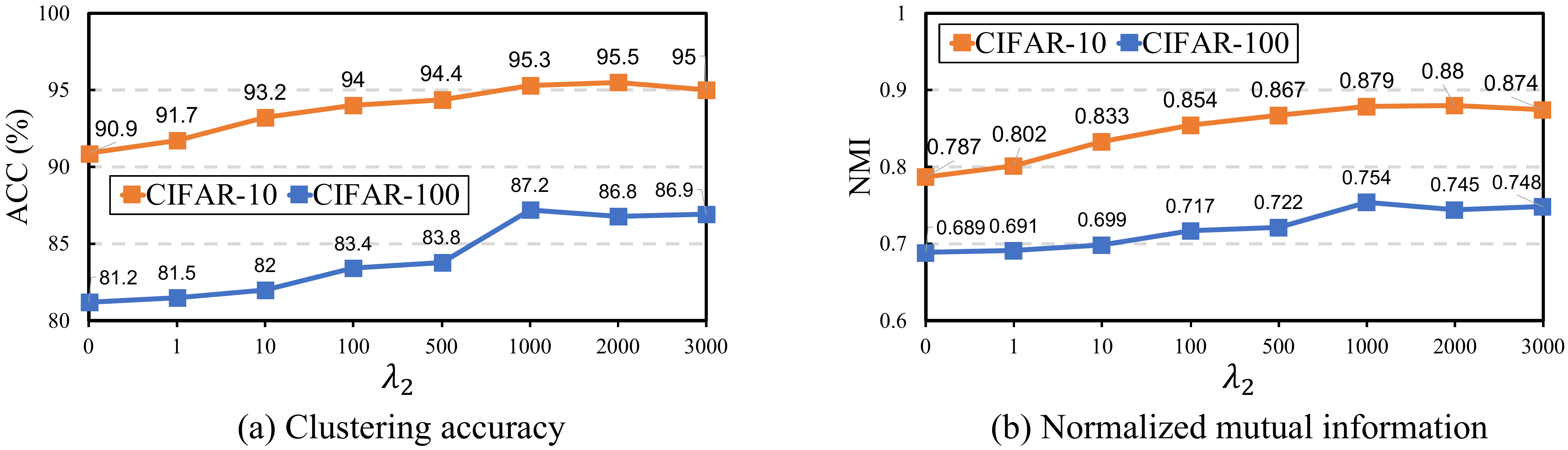}
\vspace{-.1in}
\caption{Sensitivity to the weight of \ours.}
\label{fig:weight}
\vspace{-.01in}
\end{figure}

\textbf{Impact of the weight of \ours.} In Fig.~\ref{fig:weight}, we evaluate the important hyper-parameter of our method, \textit{i.e.}, the weight of \ours ($\lambda_2$). For evaluation, we keep other hyper-parameters unchanged and vary $\lambda_2$ in a range of [0, 5000]. When $\lambda_2=0$, our method reduces to the baseline model. When inserting \ours into the system ($\lambda_2 \ge 1$), the results are consistently improved in all values. Specifically, the performance first increases with  $\lambda_2$ and becomes stable when $\lambda_2 \ge 1000$. The best results are produced when $\lambda_2$ in the range of [1000, 3000]. This indicates that our method is insensitive to the changing of $\lambda_2$ in a wide range.

\subsection{Comparison with State-of-The-Art}

\begin{table}[!t]
\caption{Comparison with state-of-the-art methods on CIFAR-10, CIFAR-100 and ImageNet for novel class discovery. Note that, RS \cite{han2020automatically} did not evaluate the NMI metric and did not provide results of 10-class setting on CIFAR-100.}
\vspace{-.2in}
\footnotesize
\begin{center}
\newcolumntype{C}{>{\centering\arraybackslash}X}%
\newcolumntype{R}{>{\raggedleft\arraybackslash}X}%
\begin{tabularx}{0.92\linewidth}{l|c|CC|cC|CC}
\hline
\multirow{2}{*}{Method} & \multirow{2}{*}{Venue} & \multicolumn{2}{c|}{CIFAR-10} & \multicolumn{2}{c|}{CIFAR-100} & \multicolumn{2}{c}{ImageNet} \\
 &  & ACC & NMI & ACC & NMI & ACC & NMI \\
\hline
K-means \cite{macqueen1967_kmeans} & Classic& 65.5\% & 0.422& 66.2\% & 0.555& 71.9\% & 0.713 \\
KCL \cite{hsu2017learning}& ICLR18 &66.5\% & 0.438 & 27.4\% & 0.151 & 73.8\% & 0.750 \\
MCL \cite{hsu2019multi}& ICLR19 &64.2\% & 0.398 & 32.7\% & 0.202 & 74.4\% & 0.762 \\
DTC \cite{han2019learning}& ICCV19 &87.5\% & 0.735 & 72.8\% & 0.634& 78.3\% & 0.791 \\
RS \cite{han2020automatically}& ICLR20 &91.7\%& - & - & - & 82.5\% & - \\
\hline
Baseline & This & 90.9\% & 0.787 & 81.2\% &0.689& 77.1\% & 0.784\\
\bf Ours & Work & \bf 95.3\% & \bf 0.879 & \bf 87.2\%& \bf 0.754 & \bf 85.7\% & \bf 0.827 \\
\hline
\end{tabularx}
\end{center}
\vspace{-.1in}
\label{tabel:state-of-the-art}
\end{table}

We compare the proposed method with state-of-the-art in novel class discovery, including: K-means \cite{macqueen1967_kmeans}, KCL \cite{hsu2017learning}, MCL \cite{hsu2019multi}, DTC \cite{han2019learning} and RS \cite{han2020automatically}. For K-means \cite{macqueen1967_kmeans}, we train the model on the labeled data and extract the last layer of the CNN as the features of unlabeled samples. Then, we directly perform K-means on unlabeled data to obtain clustering results. Compared results are shown in Table~\ref{tabel:state-of-the-art}. Our baseline achieves very competitive clustering performance compared with the state of the art. The baseline is higher than DTC \cite{han2019learning} on CIFAR-10 and CIFAR-100, and slightly lower than DTC \cite{han2019learning} on ImageNet. In addition, it is clear that our method (``Baseline+\ours'') outperforms state-of-the-art methods by a large margin. Specifically, our approach achieves \textbf{95.3\% for CIFAR-10, 90.1\% for CIFAR-100 and 85.7\% for ImageNet} in clustering accuracy. Both KCL \cite{hsu2017learning} and MCL \cite{hsu2019multi} use pairwise similarity for clustering learning. However, these two method fail to produce competitive performance. For example, KCL \cite{hsu2017learning} and MCL \cite{hsu2019multi} have similar resutls to K-means on CIFAR-10 and ImageNet, but are largely inferior to K-means on CIFAR-100. Our method produces significant superiority over KCL \cite{hsu2017learning} and MCL \cite{hsu2019multi}. Compared to DTC \cite{han2019learning}, our method surpasses it in all three datasets, especially in CIFAR-100. RS \cite{han2020automatically} is the latest method, which also uses the labeled data during unsupervised clustering. However, RS mainly focuses on using labeled data to maintain the accuracy in old classes. Compared to RS \cite{han2020automatically}, our method outperforms it by 3.6\% and 3.2\% in clustering accuracy for CIFAR-10 and ImageNet respectively, indicating that our method produces a new state of the art.

\begin{figure}[!t]
\centering
\includegraphics[width=0.9\linewidth]{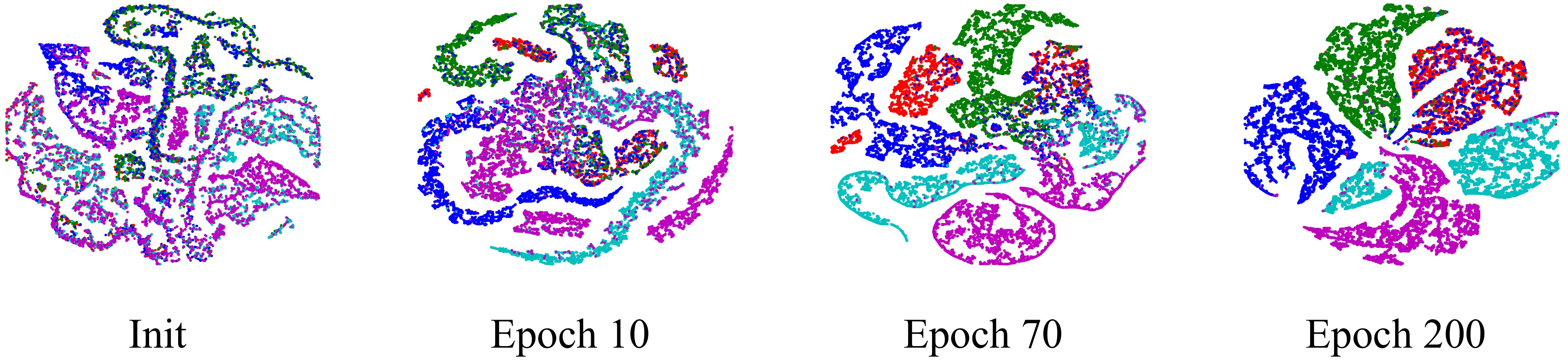}
\vspace{-.1in}
\caption{T-SNE \cite{TSNE} visualization of unlabeled samples of CIFAR-10. Results in different stages of our method are shown. Different colors represent different classes. Our method progressively separates the unlabeled samples into discriminative clusters.}
\label{fig:tsne}
\vspace{-.01in}
\end{figure}

\subsection{Visualization}

To better reflect the effectiveness of the proposed method, we visualize the distributions of unlabeled samples in different stages. Specifically, we extract the outputs of the new classifier as features of samples and map them into 2-D vectors by t-SNE \cite{TSNE}. Visualization results are shown in Fig.~\ref{fig:tsne}. In initialization, unlabeled samples are scattered. Through the training of our method, the samples of the same class are progressively grouped together, thereby enabling the new classifier to accurately recognize new classes in unlabeled data.

\section{Conclusions}

This work studies the problem of discovering novel classes in unlabeled data given labeled data of disjoint classes. To address this problem, we focus on effectively incorporating the labeled data into the step of unsupervised clustering in unlabeled data. To achieve this goal, we  present \ours to dynamically incorporate labeled samples of known classes and unlabeled samples of novel classes as well as their labels and pseudo-labels. \ours can generate joint-class samples with reliable pseudo-labels and diverse smooth samples of new classes. Learning with these generated samples help to improve the model performance of new class recognition. Experiments conducted on three image classification benchmarks demonstrate that \ours can consistently improve the performance of a competitive baseline, enabling us to achieve state-of-the-art results. 

\bibliographystyle{splncs04}
\bibliography{egbib}

\begin{thebibliography}{10}
\providecommand{\url}[1]{\texttt{#1}}
\providecommand{\urlprefix}{URL }
\providecommand{\doi}[1]{https://doi.org/#1}

\bibitem{berthelot2020remixmatch}
Berthelot, D., Carlini, N., Cubuk, E.D., Kurakin, A., Sohn, K., Zhang, H.,
  Raffel, C.: Remixmatch: Semi-supervised learning with distribution matching
  and augmentation anchoring. In: Proc. ICLR (2020)

\bibitem{berthelot2019mixmatch}
Berthelot, D., Carlini, N., Goodfellow, I., Papernot, N., Oliver, A., Raffel,
  C.A.: Mixmatch: A holistic approach to semi-supervised learning. In: Proc.
  NeurIPS (2019)

\bibitem{Chang_2017_ICCV}
Chang, J., Wang, L., Meng, G., Xiang, S., Pan, C.: Deep adaptive image
  clustering. In: Proc. ICCV (2017)

\bibitem{deng2009imagenet}
Deng, J., Dong, W., Socher, R., Li, L.J., Li, K., Fei-Fei, L.: Imagenet: A
  large-scale hierarchical image database. In: Proc. CVPR (2009)

\bibitem{ghasedi2017deep}
Ghasedi~Dizaji, K., Herandi, A., Deng, C., Cai, W., Huang, H.: Deep clustering
  via joint convolutional autoencoder embedding and relative entropy
  minimization. In: Proc. ICCV (2017)

\bibitem{han2020automatically}
Han, K., Rebuffi, S.A., Ehrhardt, S., Vedaldi, A., Zisserman, A.: Automatically
  discovering and learning new visual categories with ranking statistics. In:
  ICLR (2020)

\bibitem{han2019learning}
Han, K., Vedaldi, A., Zisserman, A.: Learning to discover novel visual
  categories via deep transfer clustering. In: Proceedings of the IEEE
  International Conference on Computer Vision. pp. 8401--8409 (2019)

\bibitem{resnet}
He, K., Zhang, X., Ren, S., Sun, J.: Deep residual learning for image
  recognition. In: Proc. CVPR (2016)

\bibitem{hsu2017learning}
Hsu, Y.C., Lv, Z., Kira, Z.: Learning to cluster in order to transfer across
  domains and tasks. In: Proc. ICLR (2018)

\bibitem{hsu2019multi}
Hsu, Y.C., Lv, Z., Schlosser, J., Odom, P., Kira, Z.: Multi-class
  classification without multi-class labels. In: Proc. ICLR (2019)

\bibitem{krizhevsky2009learning}
Krizhevsky, A., Hinton, G., et~al.: Learning multiple layers of features from
  tiny images  (2009)

\bibitem{krizhevsky2012alexnet}
Krizhevsky, A., Sutskever, I., Hinton, G.E.: Imagenet classification with deep
  convolutional neural networks. In: Proc. NeurIPS (2012)

\bibitem{macqueen1967_kmeans}
MacQueen, J., et~al.: Some methods for classification and analysis of
  multivariate observations. In: Proc. Berkeley symposium on mathematical
  statistics and probability (1967)

\bibitem{oliver2018realistic}
Oliver, A., Odena, A., Raffel, C.A., Cubuk, E.D., Goodfellow, I.: Realistic
  evaluation of deep semi-supervised learning algorithms. In: Proc. NeurIPS
  (2018)

\bibitem{pan2009survey}
Pan, S.J., Yang, Q.: A survey on transfer learning. IEEE Transactions on
  knowledge and data engineering  (2009)

\bibitem{vgg}
Simonyan, K., Zisserman, A.: Very deep convolutional networks for large-scale
  image recognition. In: Proc. ICLR (2015)

\bibitem{strehl2002cluster}
Strehl, A., Ghosh, J.: Cluster ensembles---a knowledge reuse framework for
  combining multiple partitions. JMLR  (2002)

\bibitem{tan2018survey}
Tan, C., Sun, F., Kong, T., Zhang, W., Yang, C., Liu, C.: A survey on deep
  transfer learning. In: Proc. ICANN (2018)

\bibitem{tarvainen2017mean}
Tarvainen, A., Valpola, H.: Mean teachers are better role models:
  Weight-averaged consistency targets improve semi-supervised deep learning
  results. In: Proc. NeurIPS (2017)

\bibitem{TSNE}
Van Der~Maaten, L.: Accelerating t-sne using tree-based algorithms. JMLR
  (2014)

\bibitem{vinyals2016matching}
Vinyals, O., Blundell, C., Lillicrap, T., Wierstra, D., et~al.: Matching
  networks for one shot learning. In: Proc. NeurIPS (2016)

\bibitem{weiss2016survey}
Weiss, K., Khoshgoftaar, T.M., Wang, D.: A survey of transfer learning. Journal
  of Big data  (2016)

\bibitem{wu2019deep}
Wu, J., Long, K., Wang, F., Qian, C., Li, C., Lin, Z., Zha, H.: Deep
  comprehensive correlation mining for image clustering. In: Proc. ICCV (2019)

\bibitem{xie2016DEC}
Xie, J., Girshick, R., Farhadi, A.: Unsupervised deep embedding for clustering
  analysis. In: Proc. ICML (2016)

\bibitem{yang2017towards}
Yang, B., Fu, X., Sidiropoulos, N.D., Hong, M.: Towards k-means-friendly
  spaces: Simultaneous deep learning and clustering. In: Proc. ICML (2017)

\bibitem{yun2019cutmix}
Yun, S., Han, D., Oh, S.J., Chun, S., Choe, J., Yoo, Y.: Cutmix: Regularization
  strategy to train strong classifiers with localizable features. In: Proc.
  ICCV (2019)

\bibitem{zelnik2005self}
Zelnik-Manor, L., Perona, P.: Self-tuning spectral clustering. In: Proc.
  NeurIPS (2005)

\bibitem{zhang2018MixUp}
Zhang, H., Cisse, M., Dauphin, Y.N., Lopez-Paz, D.: mixup: Beyond empirical
  risk minimization. In: ICLR (2018)

\end{thebibliography}
\end{document}